\documentclass{article}

\usepackage{titling}
\usepackage{fancyvrb}

\predate{}
\date{}
\postdate{}

\begin{document}

\title{Neuroevolution with Perceptron Turing Machines\protect\footnote{Patent pending.}}
\author{David Landaeta \protect\\ Natural Computation LLC \protect\\
\tt natucomp@gmail.com}
\maketitle

\newcommand{\qzero}{\mbox{{\bf 0}}}
\newcommand{\qone}{\mbox{{\bf 1}}}
\newcommand{\qa}{\mbox{{\bf a}}}
\newcommand{\qb}{\mbox{{\bf b}}}
\newcommand{\qc}{\mbox{{\bf c}}}
\newcommand{\qx}{\mbox{{\bf x}}}
\newcommand{\qy}{\mbox{{\bf y}}}
\newcommand{\qnn}{\mbox{{\bf n}}}
\newcommand{\qA}{\mbox{{\bf A}}}
\newcommand{\qB}{\mbox{{\bf B}}}
\newcommand{\qunk}{\mbox{{\tt ?}}}
\newcommand{\qdot}{\odot}
\newcommand{\qcross}{\otimes}
\newcommand{\qmap}[3]{\mbox{${#1}:{#2}\rightarrow{#3}$}}
\newcommand{\qangle}[1]{\mbox{$\langle{#1}\rangle$}}

\newcommand{\qmod}{\mbox{mod}}
\newcommand{\qfa}{finite automaton}
\newcommand{\qfas}{finite automata}

\newcommand{\qpred}{{\cal P}}
\newcommand{\qbool}{{\cal B}}
\newcommand{\qstr}{{\qbool^*}}
\newcommand{\qtrit}{{\cal T}}
\newcommand{\qtstr}{{\qtrit^*}}
\newcommand{\qinfstr}{\mbox{$\qbool^\infty$}}
\newcommand{\qnf}{\mbox{$\varepsilon$}}
\newcommand{\qd}{\mbox{$\delta$}}
\newcommand{\qe}{\mbox{$\epsilon$}}
\newcommand{\qs}{\mbox{$\sigma$}}
\newcommand{\qS}{\mbox{$\Sigma$}}
\newcommand{\qz}{{\cal Z}}
\newcommand{\qn}{{\cal N}}
\newcommand{\qq}{{\cal Q}}
\newcommand{\qr}{{\cal R}}
\newcommand{\qcap}{\mbox{\tt ELIN}}
\newcommand{\qCAP}{\mbox{\tt ELIN}}
\newcommand{\qnc}{\mbox{\tt NC}}
\newcommand{\qNC}{\mbox{\tt NC}}
\newcommand{\qxc}{\mbox{\tt XC}}
\newcommand{\qXC}{\mbox{\tt XC}}
\newcommand{\qdc}{\mbox{\tt DC}}
\newcommand{\qP}{\mbox{\tt P}}
\newcommand{\qPSPACE}{\mbox{\tt PSPACE}}
\newcommand{\qbox}{{\bf B}}
\newcommand{\qst}{:}
\newcommand{\qread}{\mbox{read}}

\newcommand{\qblank}{\mbox{$\sqcup$}}
\newcommand{\qbottom}{\mbox{$\perp$}}

\newcommand{\qi}[1]{{\bf #1}\index{#1}}
\newcommand{\qdd}[1]{$#1$-dimen\-sion\-al}
\newcommand{\qdD}[1]{\mbox{$#1$-D}}
\newcommand{\qext}[2]{\mbox{$#1^{(#2)}$}}
\newcommand{\qct}[2]{\mbox{$\mbox{\tt CT}^{#1}(#2)$}}
\newcommand{\qcs}[2]{\mbox{$\mbox{\tt CS}^{#1}(#2)$}}
\newcommand{\qhc}[2]{\mbox{$\underline{#1}^{#2}$}}
\newcommand{\qset}[1]{\mbox{$\underline{#1}$}}
\newcommand{\qbi}[1]{\mbox{$\overline{#1}$}}

\newcommand{\qvon}{von~Neumann}
\newcommand{\qwlog}{without loss of generality}
\newcommand{\qiff}{if and only if}
\newcommand{\qf}{\mbox{$\phi$}}
\newcommand{\qp}[1]{\mbox{$\psi_{#1}$}}
\newcommand{\qm}[1]{\mbox{$#1$}}
\newcommand{\qo}[1]{\mbox{$O(#1)$}}
\newcommand{\qtup}[1]{\mbox{$#1$-tuple}}
\newcommand{\qcua}{\mbox{$A$}}
\newcommand{\qcub}{\mbox{$B$}}
\newcommand{\qcuc}{\mbox{$C$}}
\newcommand{\qcud}{\mbox{$D$}}
\newcommand{\qpl}[1]{\mbox{$#1$-plane}}
\newcommand{\qax}[1]{\mbox{$#1$-axis}}
\newcommand{\qt}[1]{\mbox{#1}}
\newcommand{\qprop}{Property}
\newcommand{\qin}{\mbox{input}}
\newcommand{\qout}{\mbox{output}}
\newcommand{\qpi}{\mbox{$\pi$}}
\newcommand{\qipi}{\mbox{$\pi^{-1}$}}
\newcommand{\qalg}{Algorithm}
\newcommand{\qtau}[1]{\mbox{$\tau(#1)$}}
\newcommand{\qchi}{\mbox{$\theta$}}
\newcommand{\qea}{\mbox{$\protect\underline{a}$}}
\newcommand{\qeb}{\mbox{$\protect\underline{b}$}}
\newcommand{\qec}{\mbox{$\protect\underline{c}$}}
\newcommand{\qcau}{\Rightarrow^{+}}
\newcommand{\qsn}[1]{\mbox{$\#(#1)$}}
\newcommand{\qmun}{\mbox{$\mu$}}
\newcommand{\qmu}[1]{\mbox{$\mu(#1)$}}
\newcommand{\qevset}{\mbox{$E$}}
\newcommand{\qevs}[1]{\mbox{$\mbox{$\cal E$}(#1)$}}
\newcommand{\qevsi}[2]{\mbox{$\mbox{$\cal E$}(#1,#2)$}}
\newcommand{\qevn}[1]{\mbox{$n(#1)$}}
\newcommand{\qevm}[1]{\mbox{$m(#1)$}}
\newcommand{\qsncn}[1]{\mbox{$\#_{#1}$}}
\newcommand{\qsncna}{\mbox{$\#_{A}$}}
\newcommand{\qsncnb}{\mbox{$\#_{B}$}}
\newcommand{\qsncnc}{\mbox{$\#_{C}$}}
\newcommand{\qsncnd}{\mbox{$\#_{D}$}}
\newcommand{\qsnc}[2]{\mbox{$\#_{#1}(#2)$}}
\newcommand{\qsnca}[1]{\mbox{$\#_{A}(#1)$}}
\newcommand{\qsncb}[1]{\mbox{$\#_{B}(#1)$}}
\newcommand{\qsncc}[1]{\mbox{$\#_{C}(#1)$}}
\newcommand{\qsncd}[1]{\mbox{$\#_{D}(#1)$}}
\newcommand{\qep}[2]{\mbox{$\#_{#1}^{-1}(#2)$}}
\newcommand{\qepa}[1]{\mbox{$\#_{A}^{-1}(#1)$}}
\newcommand{\qepb}[1]{\mbox{$\#_{B}^{-1}(#1)$}}
\newcommand{\qepc}[1]{\mbox{$\#_{C}^{-1}(#1)$}}
\newcommand{\qepd}[1]{\mbox{$\#_{D}^{-1}(#1)$}}
\newcommand{\qdir}[1]{\mbox{$\theta(#1)$}}
\newcommand{\qti}{\mbox{$t_{\infty}$}}
\newcommand{\qdif}{\rhd}
\newcommand{\qindexentry}[2]{\item #1, #2}

\newcommand{\qnl}{\\ \hspace*{3pc}}

\newcommand{\qfor}[2]
{For #1 Do \begin{list}{}
{\setlength{\itemsep}{0in}
\setlength{\parsep}{0in}
\setlength{\topsep}{0in}}
\item #2 \end{list}}

\newcommand{\qlet}[1]{Let \qm{#1}}
\newcommand{\qreturn}[1]{return #1}

\newsavebox{\qboxa}
\newsavebox{\qboxb}
\newsavebox{\qboxc}
\newsavebox{\qboxd}
\newsavebox{\qboxw}
\newsavebox{\qboxx}
\newsavebox{\qboxy}
\newsavebox{\qboxz}

\newcounter{cas}

\newcounter{exa}[section]
\renewcommand{\theexa}{\thesection.\arabic{exa}}

\newcounter{def}[section]
\renewcommand{\thedef}{\thesection.\arabic{def}}

\newcounter{pro}[section]
\renewcommand{\thepro}{\thesection.\arabic{pro}}

\newcounter{lem}[section]
\renewcommand{\thelem}{\thesection.\arabic{lem}}

\newcounter{thm}[section]
\renewcommand{\thethm}{\thesection.\arabic{thm}}

\newcounter{cor}[section]
\renewcommand{\thecor}{\thesection.\arabic{cor}}

\newenvironment{qcases}{\setcounter{cas}{0}}{}

\newenvironment{case}
{\begin{description}\refstepcounter{cas}
\item[Case \thecas:]}{\end{description}}

\newenvironment{example}
{\refstepcounter{exa}
 \vspace{\baselineskip}
 \noindent
 {\bf Example \theexa:}}{\nolinebreak \hfill $\triangle$ \vspace{\baselineskip}}

\newenvironment{definition}
{\refstepcounter{def}
 \vspace{\baselineskip}
 \noindent
 {\bf Definition \thedef:}}{\nolinebreak \hfill $\Diamond$ \vspace{\baselineskip}}

\newenvironment{property}
{\refstepcounter{pro}
 \vspace{\baselineskip}
 \noindent
 {\bf Property \thepro:}}{\vspace{\baselineskip}}

\newenvironment{lemma}
{\refstepcounter{lem}
 \vspace{\baselineskip}
 \noindent
 {\bf Lemma \thelem:}}{\vspace{\baselineskip}}

\newenvironment{theorem}
{\refstepcounter{thm}
 \vspace{\baselineskip}
 \noindent
 {\bf Theorem \thethm:}}{\vspace{\baselineskip}}

\newenvironment{theorema}[1]
{\refstepcounter{thm}
 \vspace{\baselineskip}
 \noindent
 {\bf Theorem \thethm (#1):}}{\vspace{\baselineskip}}

\newenvironment{corollary}
{\refstepcounter{cor}
 \vspace{\baselineskip}
 \noindent
 {\bf Corollary \thecor:}}{\vspace{\baselineskip}}

\newenvironment{algorithm}[4]
{\qsingle
 \begin{figure}
 \caption{#1}
 \begin{tabbing}
 output: \= \kill
 axes:\>\+#2\- \\
 input:\>\+#3\- \\
 output:\>\+#4\- \\
 \\ }{\end{tabbing} \end{figure} \qdouble}

\newcommand{\qifthenelse}[3]
{{\bf if} \= \+ #1 \\
  {\bf then} \= \+ #2 \- \\
  {\bf else} \> \+ #3 \- \- }

\newcommand{\qcomment}[1]
{comment: \= \+ #1 \- \\ }

\newenvironment{block}
{\begin{list}{}
{\setlength{\itemsep}{0in}
\setlength{\leftmargin}{0in}
\setlength{\parsep}{0in}
\setlength{\topsep}{0in}}
\item Begin}{\item End \end{list}}

\newenvironment{proof}
{\noindent{\bf Proof:}}{\nolinebreak \hfill $\Box$ \vspace{\baselineskip}}

\newenvironment{proofa}[1]
{\vspace{\baselineskip}\noindent{\bf Proof #1:}}{\nolinebreak \hfill $\Box$ \vspace{\baselineskip}}

\begin{abstract}

We introduce the perceptron Turing machine and show how it can be used to create a system of neuroevolution. Advantages of this approach include automatic scaling of solutions to larger problem sizes, the ability to experiment with hand-coded solutions, and an enhanced potential for understanding evolved solutions.
Hand-coded solutions may be implemented in the low-level language of Turing machines, which is the genotype used in neuroevolution, but a high-level language called Lopro is introduced to make the job easier.

\end{abstract}

\section{Introduction} \label{sec:intro}

The current most popular method for creating an artificial neural network (ANN) relies on human ingenuity to create the network structure, and backpropagation during training to set values for weights and biases. This approach is effective at specific tasks, but it is generally accepted that biological nervous systems must be doing something very different when they learn new tasks.

Neuroevolution (NE) has been proposed as an alternative (or supplemental) mechanism that is more consistent with known biological processes. NE uses some form of evolutionary algorithm to either create the network structure, or set the values of weights and biases, or do both. The motivation for NE goes beyond the attempt to better understand the biological mechanisms for learning; there are some machine learning tasks for which NE seems to be more appropriate than backpropagation, such as playing games where the fitness of an ANN is only known after a sequence of interactions~\cite{risi}.

We introduce the perceptron Turing machine (PTM), which is a variant of the well-known alternating Turing machine (ATM). We show how to create an NE system by viewing the instructions of a PTM program as the genes of a genotype whose phenotype is an ANN. This NE system simultaneously evolves both the network structure and the connection weight and bias values.
According to the NE classification given in Floreano {\em et al.}~\cite{floreano}, the genotype is {\em developmental}, since it encodes a specification for building the ANN rather than encoding the ANN directly. The developmental approach has the advantage of potentially describing a large network with a small amount of code, which is useful in data compression.

\section{Alternating Turing Machines} \label{sec:atm}

We start with a definition of an ATM that emphasizes the relationship between the ATM model and the {\em uniform circuit}\ model of parallel computation~\cite{balc2}. In the uniform circuit model, a program is given numbers $m, n$ and produces a description of a Boolean circuit having $m$ outputs and $n$ inputs. Thus, running a program is a two-phase process that creates a circuit in the {\em build phase}\ and feeds inputs to the circuit producing outputs in the {\em execution phase}. This point of view is helpful, because the PTM model simply replaces a specification for building a circuit with a specification for building an ANN. More importantly, this approach provides an automatic way for the circuit (or ANN) to scale up to larger problem sizes.

An ATM $M$ with $j$ instructions and $k$ tapes is a tuple:
\[
M = (Q, \Sigma, q_0, g, p)
\]
where:
\begin{enumerate}
\item $Q$ is the finite set of {\em states};
\item $\Sigma$ is the {\em tape alphabet};
\item $q_0 \in Q$ is the {\em initial state};
\item \qmap{g}{Q}{\{\mbox{and}, \mbox{or}, \mbox{true}, \mbox{false}, \mbox{read}, \mbox{read-inverted} \}}
is a function that maps the states of $Q$ to the {\em gate types}\ of a circuit, and
\item \qmap{p}{\{1, \ldots, j\}}{I} is the {\em program}\ of $M$, where
\[
I = Q \times \Sigma^k \times Q \times \Sigma^k \times \{R,N,L\}^k
\]
is the set of all possible {\em instructions} for $M$, and $R, N, L$ are {\em tape head movements}:
$R = \mbox{move right}$,
$N = \mbox{don't move}$, and
$L = \mbox{move left}$.
\end{enumerate}

A {\em configuration}\ of $M$ is all the information needed for an instantaneous description of the machine: the current state, the contents of all $k$ tapes, and the positions of the $k$ read-write tape heads.
An instruction $p(i)$ describes how a configuration $c_1$ may (nondeterministically) lead to another configuration $c_2$ in one time step, which is denoted by $c_1 \stackrel{M}{\rightarrow} c_2$.
Suppose
\[
p(i) = (q, (a_1, \ldots, a_k), q', (a'_1, \ldots, a'_k), (m_1, \ldots, m_k)).
\]
Then whenever the current state is $q$ and the symbols $a_1, \ldots, a_k$ are scanned on the tapes, we may change to state $q'$, write symbols $a'_1, \ldots, a'_k$ on the tapes, and move the tape heads according to $m_1, \ldots, m_k$.

Any state $q$ with $g(q) \in \{ \mbox{true}, \mbox{false} \}$ is considered to be a
{\em final state}, meaning that it is not possible to transition out of a configuration in state $q$ regardless of any instructions to the contrary in $p$. We refer to such a configuration as an {\em input configuration}.

We imagine the input as an array of bits in random access memory, where each bit is directly addressable. Some of the tapes of $M$ may be designated as {\em input index tapes}. Suppose there is exactly one such tape; then $M$ reads the input bit at position $i$ by writing $i$ in binary on the input index tape and transitioning to a state $q$ with $g(q) = \mbox{read}$. At that point, $M$ automatically transitions to a state $q'$ with $g(q') = \mbox{true}$ if the input bit is 1, or it transitions to a state $q''$ with $g(q'') = \mbox{false}$ if the input bit is 0. Replacing the gate type
$\mbox{read}$ with $\mbox{read-inverted}$ in this scenario causes $M$ to read the inverse of the input bit, so that $M$ ends up in state $q'$ with $g(q') = \mbox{false}$ if the input bit is 1, or it ends up in state $q''$ with $g(q'') = \mbox{true}$ if the input bit is 0. Note that these automatic transitions override any relevant instructions in $p$.

If there are multiple input index tapes, then they are viewed as specifying the coordinates of the input bit within a multidimensional array, but apart from that, the steps required to read an input bit (or its inverse) are the same.

There may be no designated input index tapes, in which case the input is viewed as a zero-dimensional array; in other words, the input is a single bit.

Some of the tapes of $M$ may be designated as {\em output index tapes}, which are used to create the effect of producing a multidimensional array of output bits. In order to produce the output bit with coordinates $i_1, \ldots, i_d$, $M$ is started in an initial configuration having state $q_0$, the values $i_1, \ldots, i_d$ written in binary on the respective output index tapes, all other tapes having empty contents, and all tape heads at their leftmost positions. We refer to such a configuration as an {\em output configuration}.

We view $M$ as a specification for building a circuit by identifying the configurations of $M$ with the gates of the circuit. The gate type is given by applying $g$ to the current state in the configuration. For any two configurations $c_1$ and $c_2$, an input of the gate for $c_1$ is linked to the output of the gate for $c_2$ if and only if $c_1 \stackrel{M}{\rightarrow} c_2$.
A gate with type $\mbox{true}$ has no inputs and a constant output value of true. Similarly, a gate with type $\mbox{false}$ has no inputs and a constant output value of false. A gate with a type in $\{ \mbox{and}, \mbox{or} \}$ may have any number of inputs---even zero, in which case the gate acts as if its type is $\mbox{false}$,
otherwise the gate performs the logical function indicated by its type.

Some practical considerations for the build phase need to be addressed. First, we need to put limits on the lengths of all tapes; otherwise, the build might go on forever. These limits on the input and output index tapes specify the sizes of the input and output arrays, and on other tapes they specify the amount of additional work space we expect the computation to require.
We don't require all tapes to have the same size, but we do require the sizes to be fixed. This means that the tape contents can't really be empty, so instead we use a string of zeros as the default tape contents.

The computation needs to know when the end of a tape has been reached, and we achieve this by designating one tape symbol to be an {\em endmark}. We only need one endmark, because we consider the tape to be circular, so that moving right from the endmark automatically positions the tape head at the first symbol on the tape.
We allow moving left from the endmark to access the last tape symbol in one step. We consider the tape head positioned at the endmark to be the default (leftmost) position. We ignore any tape symbol overwrites in instructions of $p$ that would change an endmark to a non-endmark or a non-endmark to an endmark.

Second, we need to eliminate the possibility of cycles in the circuit. We do this by building the circuit in depth-first order, from output configurations to input configurations, which allows us to detect cycles as we go. If adding a link between two gates would create a cycle, then we simply don't add the link. Note that this is the only reason why we defined the ATM program to be an {\em ordered}\ list of instructions: the decision on which link to eliminate in order to break a cycle might depend on the order of instructions in the program. Apart from this, an unordered set of instructions would do just as well. However, we will see below in the PTM model that it is even more important for the program to be an ordered list of instructions, since the number of occurrences of a given instruction in the program has an affect on the weight and bias values in the ANN that is produced.

Further optional resource restrictions should be considered, such as limits on the total number of gates in the circuit, the depth of the circuit, or the {\em fanout}\ of the circuit, which is the maximum number of inputs on any gate. If a resource limit is exceeded, then we can either consider it to be a non-fatal error, meaning that the build is ended, but the circuit built to that point is retained and used for execution, or we can consider it to be fatal, in which case an exception is raised.

\section{Perceptron Turing Machines} \label{sec:ptm}

Our PTM model is a variant of the ATM model that replaces its specification for building a Boolean circuit with a specification for building an ANN. This is done in a straightforward way: the identification of a configuration with a logic gate becomes an identification of a configuration with a perceptron, which we hereafter refer to as a {\em node}.

A PTM $M$ with $j$ instructions and $k$ tapes is a tuple:
\[
M = (Q, \Sigma, q_0, q_1, p)
\]
where:
\begin{enumerate}
\item $Q$ is the finite set of states;
\item $\Sigma$ is the tape alphabet;
\item $q_0 \in Q$ is the {\em output state};
\item $q_1 \in Q$ is the {\em input state}, and
\item \qmap{p}{\{1, \ldots, j\}}{I} is the program of $M$, where
\[
I = Q \times \Sigma^k \times Q \times \Sigma^k \times \{R,N,L\}^k \times \{-1,1\}^2
\]
is the set of all possible instructions for $M$.
\end{enumerate}

This definition drops the explicit gate type designation of the ATM model, because an ANN only needs to distinguish between three types of nodes: output, input, and hidden. A configuration of $M$ corresponds to an output node if its state is $q_0$; it corresponds to an input node if its state is $q_1$; otherwise, it corresponds to a hidden node.
State $q_0$ acts the same as the initial state of an ATM, with the same relationship to output index tapes, and state $q_1$ acts like an ATM state with a gate type of
$\mbox{read}$, with the same relationship to input index tapes.

An instruction $p(i)$ is interpreted the same as an ATM instruction, except that it now has an additional pair $(\Delta w, \Delta b)$ of integers taken from the set $\{-1,1\}$. We say that $\Delta w$ is the {\em differential weight}\ and $\Delta b$ is the {\em differential bias}\ associated with the instruction. For any two configurations $c_1, c_2$, if there are any instructions in $p$ that would create a link $c_1 \stackrel{M}{\rightarrow} c_2$, then we add up all the differential weights associated with those instructions in order to find the actual weight associated with the link. We find the bias value associated with the node corresponding to a configuration $c$ by adding up all differential bias values of any instructions in $p$ that would create a link of the form $c \stackrel{M}{\rightarrow} c'$ for some configuration $c'$, with the default bias being zero if no such $c'$ exists.

As an example, consider how the node for a configuration $c$ might behave like an and-gate with inputs coming from nodes for configurations $c_1$ and $c_2$.
Of course, there must exist a sub-list $p'$ of
instructions in $p$ that allow
$c \stackrel{M}{\rightarrow} c_1$ and
$c \stackrel{M}{\rightarrow} c_2$,
and there must not exist any other instructions in $p$ that would allow a link of the form $c \stackrel{M}{\rightarrow} c'$ for some configuration $c'$,
but we need to determine the differential weight and differential bias values associated with the instructions in $p'$ in order to create and-like behavior.
Assume we define the output of a perceptron in terms of its inputs using the following standard function.
\[
f({\bf x}) = \left\{ \begin{array}{rl}
		1 & \mbox{if ${\bf w \cdot x} + b > 0$} \\
		0 & \mbox{otherwise}
	\end{array}
\right.
\]
where ${\bf x}$ is the vector of real-valued inputs,
${\bf w}$ is the vector of real-valued weights,
$b$ is the real-valued bias, and
${\bf w \cdot x}$ denotes the dot product.
Then, to create and-like behavior, it suffices to have exactly four instructions in $p'$, two of which allow $c \stackrel{M}{\rightarrow} c_1$, and the other two allowing
$c \stackrel{M}{\rightarrow} c_2$, where all four have a differential weight of $1$,
and exactly three of the four have a differential bias of $-1$.

\section{Genetic Operators} \label{sec:genops}

The program $p$ of a PTM $M$ can be used as the genotype for a system of neuroevolution, where the instructions in $p$ are the genes. Since $p$ is both a program and a genotype, the system is not just an example of an evolutionary algorithm, but it is also an instance of genetic programming (GP). Moreover, the programming language used in this GP system is clearly {\em Turing complete}~\cite{banzhaf}, because the PTM model is a generalization of the Turing machine model.

But what would be wrong with using the ATM model for our GP genotype rather than the PTM model?
The big problem with this approach is that it becomes almost impossible for
evolution, starting from a random population, to converge to anything but a
trivial program---one that ignores its input and produces constant output.
For, in order for an ATM program to be non-trivial, it must have at
least one instruction that applies to an output configuration. Because such
instructions are necessary for high fitness, they quickly spread throughout
the population, causing the typical member to have several distinct
instructions that apply to the output configuration. Thus, the output
configuration typically corresponds to a gate with several inputs.
But the output configuration must represent either an and-gate or an
or-gate. If it is an and-gate with several inputs, then it is extremely likely
to always output false. Similarly, an or-gate with several inputs is extremely
likely to always output true. As a result, it is almost a certainty that the
population will converge to a trivial program regardless of how training is
performed.

By contrast, an output configuration in the PTM model corresponds to a perceptron,
which is just as likely to output false as it is to output true, assuming the population starts out randomly. Moreover, the PTM model is much more {\em robust}\ under genetics than the ATM model, meaning that the phenotype is not likely to be radically altered
when the normal genetic operators are applied to the genotype.
This is due to the fact that a PTM instruction only modifies the network weights and biases in a small incremental fashion.
Rather than, say, an and-gate changing to an or-gate in a single generation using ATM, we might have a perceptron smoothly changing from and-like behavior to or-like behavior over many generations using PTM.

In order to fully leverage this property of the PTM genotype, we should use at least one genetic operator that reorders genes within a genotype. For example, an {\em inversion operator}~\cite{goldberg} would satisfy this requirement. When combined with crossover and a fine-grained mutation operator---one that can change only an individual component of an instruction, like the differential weight or differential bias---this creates the effect of smoothly searching the space of all weights and biases between nodes.

\section{Lopro: A High-level Language for PTM Programming} \label{sec:lopro}

One of the advantages of using the PTM model for neuroevolution is that it is actually an instance of GP. This makes the evolved genotype more than just a black box---it is a human-readable computer program. Granted, the programming language is very low-level, but it is based on the well-established language for programming Turing machines, so there is reasonable hope that an expert will be able to understand how the evolved ANN makes decisions.

But we can also turn this around: the fact that this is GP means that an expert can hand-code a genotype to implement a known solution to a problem. This has multiple advantages, including:
\begin{enumerate}
\item It allows the creation of {\em sanity checks}\ on our neuroevolution system: we can test using a problem with a known solution, and seed the initial population of genotypes with that known solution. If our process of evolution destroys the solution, then we know we have a bug somewhere in the system.
\item It allows us to seed the initial population with one or more known {\em partial solutions}\ to a target problem, in the hope of giving evolution a head start on producing a better solution.
\item Even in cases where we have no idea what the solution is, we can use our experience with hand-coded PTM programs to estimate the resource requirements---number of states, number of tapes, etc.---for an evolved solution to the target problem.
\end{enumerate}

These tasks are made easier by the fact that we can create a high-level programming language---which doesn't require an expert on Turing machines to understand---and implement a translator to change the high-level code into our low-level PTM code. We have done exactly that, and we call our high-level language Lopro, which is an acronym for ``logic programming.'' It uses the fact that the ATM (and PTM) model is very good at describing solutions in terms of first order logic.

\begin{figure}
\caption{Determine if any input bit is set}
\label{fig:exists}
\begin{Verbatim}[fontsize=\small]
#include <bitset>
#include <iostream>
#include "lopro.h"

lopro::State Exists(lopro::Tape test_tape, lopro::State test_state) {

    lopro::Machine machine = test_tape.machine();
    lopro::Scope scope(machine);
    lopro::Head test_head = test_tape.head();
    lopro::State output = machine.NewState();
    lopro::State choice = machine.NewState();
    lopro::State test_end = machine.NewState();

    output =
        choice AFTER { ++test_head; };
    choice WHEN (test_head.is_end()) =
        test_end;
    choice =
        choice AFTER { *test_head++ = 0; }
        or
        choice AFTER { *test_head++ = 1; };
    test_end WHEN (test_tape >= test_tape.end()) =
        false;
    test_end =
        test_state;

    return output;
}

void main() {
    const int num_input_bits = 6;
    lopro::Machine machine;
    lopro::Tape input_index = machine.NewTape(num_input_bits);
    lopro::State input_state = machine.NewInputState(input_index);
    lopro::State output_state = machine.NewOutputState();
    output_state = Exists(input_index, input_state);
    machine.Build();
    std::bitset<num_input_bits> input("001101");
    std::bitset<1> output;
    machine.Execute(input, &output);
    std::cout << "The output is " << output[0] << std::endl;
}
\end{Verbatim}
\end{figure}

\begin{figure}
\caption{Find the transitive closure of a given directed graph}
\label{fig:trans}
\begin{Verbatim}[fontsize=\small]
#include <bitset>
#include <fstream>
#include "lopro.h"

lopro::State TransitiveClosure(lopro::State edge_x_to_y, lopro::Tape x, lopro::Tape y) {

    lopro::Machine machine = edge_x_to_y.machine();
    lopro::Scope scope(machine);
    lopro::Tape z = machine.NewTape(x.end());
    lopro::Tape path_length = machine.NewTape(x.end() * 2);
    lopro::Head depth = path_length.head();
    lopro::State output = machine.NewState();
    lopro::State path_x_to_y = machine.NewState();
    lopro::State path_x_to_y_thru_z = machine.NewState();

    output =
        path_x_to_y AFTER { ++depth; };
    path_x_to_y WHEN (depth.is_end()) =
        edge_x_to_y;
    path_x_to_y =
        path_x_to_y AFTER { ++depth; }
        or
        Exists(z, path_x_to_y_thru_z); // from previous example
    path_x_to_y_thru_z =
        path_x_to_y AFTER { y = z; z = 0; ++depth; }
        and
        path_x_to_y AFTER { x = z; z = 0; ++depth; };

    return output;
}

void main() {
    const int num_vertices = 4;
    lopro::Machine machine;
    lopro::Tape row_index = machine.NewTape(num_vertices);
    lopro::Tape col_index = machine.NewTape(num_vertices);
    lopro::State input_state = machine.NewInputState(row_index, col_index);
    lopro::State output_state = machine.NewOutputState(row_index, col_index);
    output_state = TransitiveClosure(input_state, row_index, col_index);
    machine.Build();
    std::ifstream in("adjacency_matrix.txt");
    std::ofstream out("transitive_closure.txt");
    std::bitset<num_vertices * num_vertices> input;
    std::bitset<num_vertices * num_vertices> output;
    in >> input;
    machine.Execute(input, &output);
    out << output;
}
\end{Verbatim}
\end{figure}

Lopro is currently implemented as a collection of class, function, and macro definitions in C++ following the C++11 standard. It leverages powerful features of C++, including operator overloading and lambda expressions, in order to simplify the notation. Figure~\ref{fig:exists} shows a simple example program, which determines if there exists any bit in the input that is set to true. A more complex example is given by Figure~\ref{fig:trans}, which computes the transitive closure of a given directed graph, where the input and output graphs are specified by their adjacency matrices.

The heart of Lopro is the Machine class, which embodies the PTM itself. It is a container for Tape and State objects as well as instructions. A Machine starts out in {\em definition mode}, during which tapes, states, and instructions are added to it.
The various ``New...'' methods add tapes or states and return objects that are used to refer to what was just added. All such Lopro objects wrap a smart pointer to the originating Machine plus a unique integer identifier for the object, so it is always safe and efficient to copy such objects by value.
The Build method ends definition mode, builds the ANN according to the definition, and puts the Machine in {\em execution mode}, during which we can feed inputs and receive outputs from the built ANN.

Every Tape object has an immutable ``end'' property, which is the integer specified in the argument to the NewTape method of Machine. This is how we limit the size of the tape, but rather than directly specifying the number of bits the tape can hold, the end value is one greater than the largest unsigned integer that we expect to be contained on the tape in binary. This is because the tape typically contains an array index, so it is normally most convenient to express the end value as the size of the corresponding array. Thus, the NewTape method computes the appropriate limit on the number of bits the tape can hold based on the specified end value. However, we must be aware that, because we allow any bit on the tape to be overwritten using the tape's Head object, it may be possible (if the end value is not an exact power of two) to write a value on the tape that is greater than or equal to the end value. Fortunately, this condition is easy to detect and correct, as can be seen in the Exists function in Figure~\ref{fig:exists}.

The input state is added with the NewInputState method of Machine, which takes a variable-length argument list denoting the input index tapes. Similarly, the NewOutputState method adds the output state and specifies the output index tapes. All other states are added with the NewState method, which takes no arguments.

The assignment operator is overloaded in the State class to provide a convenient way of creating instructions. An instruction in Lopro is a high-level analog of a PTM instruction. Like the latter, a Lopro instruction has a {\em precondition}\ on the machine configuration that must be satisfied in order for the instruction to apply, and it has an {\em action}\ that changes the configuration when the instruction is applied. Using the overloaded assignment operator, an instruction has the general form:
\begin{center}
\begin{BVerbatim}
precondition_expression = action_expression;
\end{BVerbatim}
\end{center}
The precondition expression may be just a State object, meaning that the precondition is satisfied whenever the machine is in that state regardless of tape contents or head positions. We would use a {\em when clause}---a usage of the WHEN macro---in order to add constraints on the tape contents or head positions, like so:
\begin{center}
\begin{BVerbatim}
from_state WHEN (conditional_expression) = action_expression;
\end{BVerbatim}
\end{center}
The conditional expression intensionally has the appearance of a Boolean expression in terms of Tape and Head objects, but the actual result type of the expression is an internal Conditional class, which can only be used in a when clause.

We can reproduce exactly the low-level PTM preconditions on tape contents and head positions by using the Head object, which is accessed using the immutable ``head'' property of the Tape object. Operators are overloaded in the Head class so that it behaves like an iterator for an array of bits. For example, the conditional expression:
\begin{center}
\begin{BVerbatim}
*tape_head == 1
\end{BVerbatim}
\end{center}
is satisfied whenever the Head object ${\tt tape\_head}$ is scanning the symbol 1. The accessor method ${\tt is\_end}$ of the Head class is used to determine if the head is currently scanning the endmark:
\begin{center}
\begin{BVerbatim}
tape_head.is_end()
\end{BVerbatim}
\end{center}
Operators have been overloaded in the Conditional class so that such objects behave like Boolean values. Thus, the conditional expression:
\begin{center}
\begin{BVerbatim}
*tape_head == 1 or tape_head.is_end()
\end{BVerbatim}
\end{center}
is satisfied whenever the Head object ${\tt tape\_head}$ is scanning either the symbol 1 or the endmark.

On the other hand, we can create high-level preconditions by using the fact that operators have been overloaded in the Tape class so that Tape objects behave like unsigned integers. For example, in the Exists function of Figure~\ref{fig:exists}, we see the conditional expression:
\begin{center}
\begin{BVerbatim}
test_tape >= test_tape.end()
\end{BVerbatim}
\end{center}
which is satisfied whenever the contents of ${\tt test\_tape}$ is greater than or equal to the end value of the tape. We use the convention that the left-most bits on the tape---accessed by moving the head right from the endmark---are the low-order bits of the unsigned integer corresponding to the tape contents.

If multiple instructions have a precondition expression mentioning the same State object $s$, then, for any given configuration $c$ having state $s$, only the first such instruction whose precondition is satisfied by $c$ will have its action applied to $c$. The last such instruction typically has no when clause, so that it acts as a catch-all. It is obviously a good idea to group all such instructions together---effectively creating a single if-then-else statement.

The action expression of an instruction may be just a State object, meaning the configuration is changed to the specified state, but there is no change to tape contents or head positions. We would use an {\em after clause}---a usage of the AFTER macro---in order to specify changes to tape contents or head positions, like so:
\begin{center}
\begin{BVerbatim}
precondition_expression = to_state AFTER { action_statements... };
\end{BVerbatim}
\end{center}
The action statements may be low-level---using Head objects as iterators, or high-level---using Tape objects as unsigned integers, or it may be some combination of the two. Incrementing a Head object has the effect of moving the head right, and decrementing the Head object moves the head left.

Operators are overloaded for action expressions so that they behave like Boolean expressions. This gives us a high-level mechanism for specifying the differential weight and differential bias values of the corresponding PTM instructions. For example, the Lopro instruction:
\begin{center}
\begin{BVerbatim}
from_state = to_state_1 and to_state_2;
\end{BVerbatim}
\end{center}
translates to two PTM instructions $r_1$ and $r_2$, creating network links of the form
$c \stackrel{M}{\rightarrow} c_1$ and $c \stackrel{M}{\rightarrow} c_2$,
respectively, such that configuration $c$ (in state ${\tt from\_state}$) corresponds to a perceptron behaving like an and-gate with inputs coming from $c_1$ (in state ${\tt to\_state\_1}$) and $c_2$ (in state ${\tt to\_state\_2}$).

A subtlety of Lopro instruction definition is that the conditional expressions and action statements are not actually executed until the Build method is invoked. The WHEN and AFTER macros are hiding lambda expressions (capturing all variables by value) that make this possible. This is notationally very convenient, but it presents a technical problem for the use of Lopro objects as local variables within sub-functions: we can't rely on the constructors and destructors of these local objects to provide appropriately timed initialization and cleanup, since the associated instructions are invoked outside of the lifetimes of the local variables.

This is the motivation for the Scope class. Every Machine object maintains a stack of Scope objects, which initially contains a default scope representing the main function. A Scope instantiation in a sub-function like the following:
\begin{center}
\begin{BVerbatim}
lopro::Scope scope(machine);
\end{BVerbatim}
\end{center}
pushes ${\tt scope}$ onto the top of the stack maintained by ${\tt machine}$, and the corresponding destructor call for ${\tt scope}$ pops it off the stack. Whenever any ``New...'' method is invoked, and whenever the ``head'' property of a Tape is accessed, the Machine records that event and associates it with the Scope object at the top of the stack. During the build phase, the Machine can then detect when an action is taking place that changes the scope, and in that case it automatically invokes appropriate initialization or cleanup operations on the affected Lopro objects. The net result is that we can always rely on the following when we define instructions:
\begin{enumerate}
\item Tape contents initially have all bits set to 0, except for output index tapes, which are initialized to the appropriate coordinate of the output bit array.
\item Heads are initially scanning the endmark.
\end{enumerate}
In addition, Tape objects associated with a sub-function scope are {\em recycled}\ if at all possible, which decreases the number of tapes used by the machine. This is important, because the maximum number of nodes in the ANN increases exponentially with the number of tapes.

There are two very desirable properties of Lopro programming. The first is that we can treat State objects as if they are {\em predicates}\ in a system of first order logic in which the variables stand for Tape objects, which can be treated as unsigned integers. From this point of view, a Lopro program simply defines an output predicate logically in terms of an input predicate. As an added bonus, this interpretation applies to sub-functions as well, resulting in a modular system of logic programming.

Note that the Exists function of Figure~\ref{fig:exists} defines the {\em existential quantifier}\ for this system of first order logic. As an exercise for the reader, write a function to define the corresponding {\em universal quantifier}. Hint: change two lines in the body of the Exists function, and rename the function to All. If you think you only need to change one line in the body, then think again.

The second desirable property of Lopro programming is that it naturally lends itself to efficient parallel computation. A good example is provided by the solution to the transitive closure problem shown in Figure~\ref{fig:trans}. This solution is essentially a proof that the transitive closure problem is in the parallel complexity class
$NC_2$, which is the class of all problems solvable by uniform circuits of size
$n^{O(1)}$ and depth $O(\log^2 n)$, and is accepted as a conservative definition for the complexity class representing efficient parallel computation~\cite{papa}.
Such solutions arise naturally from the fact that an operation that scans over a tape completely in one direction requires only logarithmic depth, and many problems, like this one, require only a logarithmic number of such operations on any path from an output node to an input node.

\section{Conclusion} \label{sec:conclusion}

Using perceptron Turing machines for neuroevolution has the following advantages.
\begin{enumerate}
\item Both the network structure and the connection weight and bias values are found through evolution.
\item A large network can be described with a small genotype.
\item Solutions automatically scale up to larger problem sizes.
\item The genotype is a human-readable computer program.
\item The programming language used by the genotype is Turing complete.
\item The genotype is robust, which allows the solution space to be explored by evolution in a smooth fashion.
\item We have the ability to experiment with hand-coded solutions in both low-level code and high-level code.
\end{enumerate}

\appendix


\begin{thebibliography}{99}

\bibitem{balc2}
Balc\'{a}zar JL, Diaz J, Gabarr\'{o} J.
{\em Structural Complexity II}, Springer-Verlag, Berlin, 1990.

\bibitem{banzhaf}
Banzhaf W, Nordin P, Keller RE, Francone FD. {\em Genetic Programming: An Introduction}, Morgan Kaufmann, San Francisco, 1998.

\bibitem{floreano}
Floreano D, D\"{u}rr P, and Mattiussi C. Neuroevolution:
from architectures to learning. {\em Evolutionary Intelligence}, 1(1):47–62, 2008.

\bibitem{goldberg}
Goldberg D. {\em Genetic Algorithms in Search, Optimization
and Machine Learning}, Addison-Wesley, Reading, MA, 1989.

\bibitem{papa}
Papadimitriou CH. {\em Computational Complexity},
Addison-Wesley, New York, 1994.

\bibitem{risi}
Risi S and Togelius J. Neuroevolution in games: State of the art and open
challenges. {\em IEEE Transactions on Computational Intelligence and AI in Games},
9(1):25–41, 2017.

\end{thebibliography}
\end{document}